\documentclass[sigconf]{acmart}
\usepackage{multirow}

\usepackage{amsmath}
\usepackage{algorithm} 
\usepackage{algpseudocode} 
\DeclareMathOperator*{\argmax}{arg\,max}
\DeclareMathOperator*{\argmin}{arg\,min}

\AtBeginDocument{%
  \providecommand\BibTeX{{%
    \normalfont B\kern-0.5em{\scshape i\kern-0.25em b}\kern-0.8em\TeX}}}

\setcopyright{acmcopyright}
\copyrightyear{2018}
\acmYear{2018}
\acmDOI{XXXXXXX.XXXXXXX}

\acmConference[Conference acronym 'XX]{Make sure to enter the correct
  conference title from your rights confirmation email}{June 03--05,
  2018}{Woodstock, NY}
\acmPrice{15.00}
\acmISBN{978-1-4503-XXXX-X/18/06}

\copyrightyear{2023}
\acmYear{2023}
\setcopyright{acmlicensed}\acmConference[KDD '23]{Proceedings of the 29th ACM SIGKDD Conference on Knowledge Discovery and Data Mining}{August 6--10, 2023}{Long Beach, CA, USA}
\acmBooktitle{Proceedings of the 29th ACM SIGKDD Conference on Knowledge Discovery and Data Mining (KDD '23), August 6--10, 2023, Long Beach, CA, USA}
\acmPrice{15.00}
\acmDOI{10.1145/3580305.3599399}
\acmISBN{979-8-4007-0103-0/23/08}

\begin{document}

\title{Knowledge Graph Reasoning over Entities and Numerical Values}

\author{Jiaxin Bai}
\authornote{~~~Work done during an internship at Amazon.}
\affiliation{%
\institution{Department of CSE, HKUST}
 \country{Hong Kong SAR, China}
}
\email{jbai@connect.ust.hk}

\author{Chen Luo}
\affiliation{%
  \institution{Amazon.com Inc}
  \city{Palo Alto}
  \country{USA}}
\email{cheluo@amazon.com}

\author{Zheng Li}
\affiliation{%
  \institution{Amazon.com Inc}
  \city{Palo Alto}
  \country{USA}}
\email{amzzhe@amazon.com}

\author{Qingyu Yin}
\affiliation{%
  \institution{Amazon.com Inc}
  \city{Palo Alto}
  \country{USA}}
\email{qingyy@amazon.com}

\author{Bing Yin}
\affiliation{%
  \institution{Amazon.com Inc}
  \city{Palo Alto}
  \country{USA}}
\email{alexbyin@amazon.com}

\author{Yangqiu Song}
\authornote{~~~Visiting academic scholar at Amazon.}
\affiliation{%
  \institution{Department of CSE, HKUST}
 \country{Hong Kong SAR, China}
}
\email{yqsong@cse.ust.hk}


\begin{abstract}

A complex logic query in a knowledge graph refers to a query expressed in logic form that conveys a complex meaning, such as \textit{where did the Canadian Turing award winner graduate from?} Knowledge graph reasoning-based applications, such as dialogue systems and interactive search engines, rely on the ability to answer complex logic queries as a fundamental task. In most knowledge graphs, edges are typically used to either describe the relationships between entities or their associated attribute values. An attribute value can be in categorical or numerical format, such as dates, years, sizes, etc. However, existing complex query answering (CQA) methods simply treat numerical values in the same way as they treat entities. This can lead to difficulties in answering certain queries, such as \textit{which Australian Pulitzer award winner is born before 1927}, and \textit{which drug is a pain reliever and has fewer side effects than Paracetamol.} In this work, inspired by the recent advances in numerical encoding and knowledge graph reasoning, we propose numerical complex query answering. In this task, we introduce new numerical variables and operations to describe queries involving numerical attribute values. To address the difference between entities and numerical values, we also propose the framework of Number Reasoning Network (NRN) for alternatively encoding entities and numerical values into separate encoding structures. During the numerical encoding process, NRN employs a parameterized density function to encode the distribution of numerical values. During the entity encoding process, NRN uses established query encoding methods for the original CQA problem. Experimental results show that NRN consistently improves various query encoding methods on three different knowledge graphs and achieves state-of-the-art results.

\end{abstract}





\maketitle

\section{Introduction}
\begin{figure*}[t]
    
  \centering
  \includegraphics[width=\linewidth]{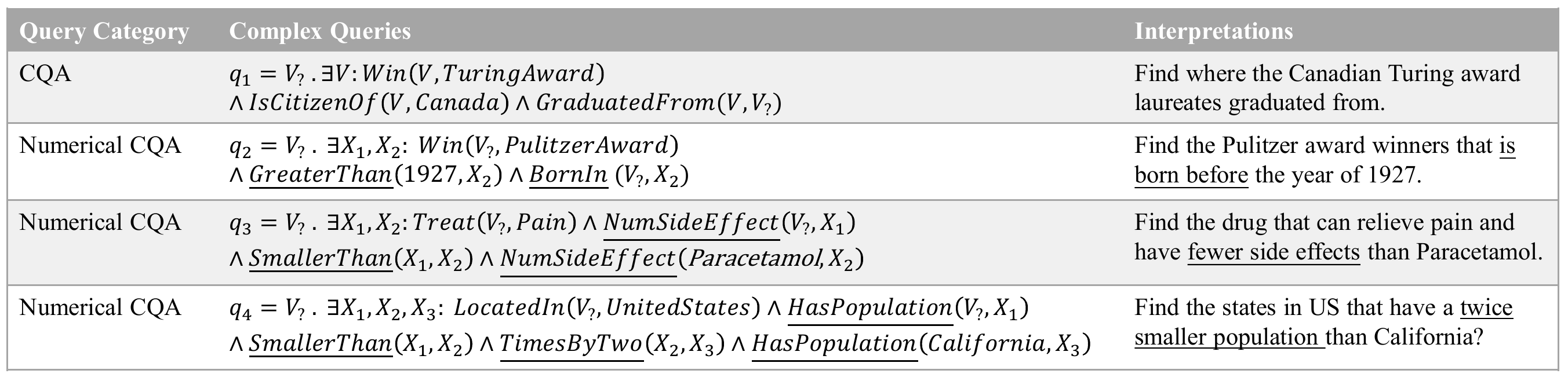}
  \vspace{-0.6cm}
  \caption{Examples of entity relations and attribute complex queries. Existing complex query answering research only focuses on entities and their relations. The queries involving numerical  values are not defined and studied in the previous literature. Here we underline the logic terms that involve at least one numerical variable in this figure.  }
  \vspace{-0.3cm}
  \label{fig:teaser}
 
\end{figure*}

Reasoning over knowledge graphs (KG) is the process of deriving new knowledge or drawing new conclusions from the existing ones in the KG~\cite{chen2020review}.
Complex query answering (CQA) is a recently developed knowledge graph reasoning task, which aims to answer complex knowledge graph queries~\cite{hamilton2018embedding,ren2020query2box,ren2020beta}.
As shown in Figure \ref{fig:teaser}, a complex knowledge graph query, or in short, complex query, targets finding the entities from the KG such that the logic expression can be satisfied~\cite{hamilton2018embedding, ren2020query2box}. 
The logic expression of a complex query usually contains multiple terms connected by logic connectives, thus it is able to carry a complicated meaning. 
For example, in Figure~\ref{fig:teaser}, 
the logic form of $q_1$ includes terms like \textit{Win(TuringAward,V)} and  operations like conjunctions $\land$ and disjunctions $\lor$, and it carries the meaning of \textit{Find where the Canadian Turing award laureates graduated from}.
The complex query answering (CQA) task is considered a KG reasoning task because it needs to deal with the incompleteness problem of the KGs.
The KGs, like Freebase~\cite{bollacker2008freebase} and YAGO~\cite{yago}, are typically sparse and incomplete. 
They often include missing relations and attribute values, although these missing edges can be inferred from other edges in the KG. 
Consequently, subgraph matching algorithms cannot directly be used to find the answers that need to be inferred from missing edges and attributes. 

Query encoding methods~\cite{hamilton2018embedding, sun2020faithful,ren2020query2box, ren2020beta, guo-kok-2021-bique, zhang2021cone, liu2021neural,bai-etal-2022-query2particles} are proposed to address this incompleteness challenge in CQA. 
In query encoding methods, queries and entities are simultaneously encoded into the same embedding space. 
Then the answers are retrieved according to the similarity between the query embedding and entity embedding.
In detail, there are two steps in encoding a complex query.
First, the query is parsed into a computational graph as a directed acyclic graph (DAG). 
Then, the query encoding is iteratively computed along the computational graph, by using neural parameterized logic operations and relational projections.
Different query encoding methods encode their queries into different structures, and they have different parameterizations for their logic and projection operations as well. 
For example, GQE~\cite{hamilton2018embedding} uses vectors, Query2Box~\cite{ren2020query2box} uses hyper-rectangles, and Query2Particles~\cite{bai-etal-2022-query2particles} uses multiple vectors in the embedding space.  

 
On the other hand, numerical values are also important components of various knowledge graphs like Freebase~\cite{bollacker2008freebase}, DBpedia~\cite{bizer2009dbpedia}, and YAGO~\cite{yago}.  
Though existing query encoding methods achieve good performance on complex query answering,  
they typically focus only on the relations between entities while being unable to reasonably deal with attribute values. 
There are in general two reasons.
First, the query encoding methods are unable to encode the semantic information of numerical values.
For $q_2$ in Figure~\ref{fig:teaser}, to find its corresponding answers, we need to first encode the semantics of \textit{the year of 1927} such that we can compute the distributions of \textit{the years before 1927}.  
However, existing query encoding methods can only learn the graph features of \textit{1927}, and ignore its numerical meaning. 
Second, the existing query encoding method is unable to describe the distributional changes of values when there are computations and comparisons between numerical values.  
In Figure~\ref{fig:teaser}, the complex queries $q_2$, $q_3$, and $q_4$ involve numerical comparisons $GreaterThan$ or $SmallerThan$.
Meanwhile, $q_4$ includes the numerical computation of $TimesByTwo$.
As the existing methods only have relational projections between entities,
they are insufficient to deal with the relations involving numerical values. 

To formally define and evaluate the problem of complex query answering with numerical values, we propose the task of numerical complex query answering (Numerical CQA). 
In Numerical CQA, we extend the existing problem scope to further include the numerical variables.
We also introduce new logic terms expressing numerical  relationships. 
In Figure~\ref{fig:teaser}, the logic terms expressing numerical attributes are underlined. 
Three example numerical queries are given in Figure~\ref{fig:teaser}, and their definitions will be given in \S\ref{section:definition}.
Meanwhile, we also create the corresponding benchmark dataset to evaluate the performance of Numerical CQA.
To do this, we use three public knowledge graphs: Freebase, YAGO, and DBpedia.

To address the challenges of reasonably encoding both entities and numerical values in a query, we propose the Number Reasoning Network (NRN)  as a solution. 
In NRN, there are two encoding phases, which are respectively used for encoding entities and numerical values, following the computation graph. 
The descriptions of NRN will be given in \S\ref{section:method}. 
Experiment results show that the NRN can constantly improve the performance on Numerical CQA when combined with different types of query encoding methods.\footnote{Experiment code available: https://github.com/HKUST-KnowComp/NRN}
The major contributions of this paper are three-fold:
\begin{itemize}
    \item We propose a new problem of numerical complex query answering (Numerical CQA) aiming at answering the complex queries that need reasoning over numerical values.
    \item We create a new benchmark dataset based on three widely used knowledge graphs: Freebase, Dbpedia, and YAGO, for evaluating the performance of Numerical CQA.
    \item We propose Number Reasoning Network (NRN) for CQA. It iteratively encodes a numerical complex query into two different structures for entities and values respectively and achieves state-of-the-art Numerical CQA. 
\end{itemize}

\begin{table}[t]
\caption{The table for frequent notations with  definitions.}
\vspace{-0.4cm}
\small
\begin{tabular}{l|l}
\toprule
Notations & Definitions \\ 
\midrule
       $\mathcal{G}$   &      The knowledge graph       \\
       $\mathcal{V}$   &    The set of knowledge graph vertices in KG  \\
      $\mathcal{R}$    &   The set of relations between entities in KG  \\
      $\mathcal{A}$    &   The set of attribute types in KG \\
      $\mathcal{N}$    &   The set of numerical values in KG \\
      $V_a$, $X_a$   &  The anchor variables that are known in a KG query\\
      $V_i$, $X_i$    & The existentially quantified variables in a KG query \\
       $r(V,V')$  & Whether the relation $r$ holds between $V$ and $V'$ \\
       $a(V,X)$  &  Whether $V$ has an attribute $a$ with value $X$ \\
       $f(X,X')$  & Whether values $X$ and $X'$ satisfy numerical relation $f$ \\
\bottomrule
\end{tabular}
\vspace{-0.4cm}
\end{table}

 \vspace{-0.1cm}
\section{Definition of Numerical CQA }
\label{section:definition}

To precisely describe the problem,
we mathematically define numerical complex query answering in this section. 
A multi-relational knowledge graph with numerical attributes is defined as $\mathcal{G}=(\mathcal{V},\mathcal{R},\mathcal{A},\mathcal{N})$. $\mathcal{V}$ is the set of the nodes or vertices, $\mathcal{R}$ is the set of relation types, $\mathcal{A}$ is the set of numerical attribute types, and $\mathcal{N}$ is the set of numerical attribute values. 
Each vertex $v \in \mathcal{V}$ represents an entity. 
Each of the relations $r \in \mathcal{R}$ is a binary function defined as $r:\mathcal{V} \times \mathcal{V} \rightarrow \{0,1\}$.
For any $r \in \mathcal{R}$, and $u,v \in \mathcal{V}$, there is a relation  $r$ between $u$ and $v$ if and only if $r(u,v)=1$.
Meanwhile, each attribute $a \in \mathcal{A}$ is also a binary function describing numerical attribute values, 
and it is defined on $ \mathcal{V}\times \mathcal{N} \rightarrow \{ 0,1 \} $. 
The entity $v \in  \mathcal{V}$ has an attribute $a \in \mathcal{A}$ of value $x \in \mathcal{N}$, if and only if  $a(v,x)=1$. 
In some queries, we also need to express the numerical relations, like comparison and computation between two numerical values, 
so we use another set of binary functions $f(x_1,x_2):\mathcal{N} \times \mathcal{N} \rightarrow \{0,1\}$ to describe whether two numerical values satisfy certain constraints, 
such as $x_1<x_2$,  $x_1>x_2+20$, or $x_1=2 x_2+3$.
 The relations are defined as binary functions so that they can be directly used to express boolean values in complex queries, and we will introduce the definition of the complex queries in the following paragraph. 

A complex numerical query is defined in existential positive first-order logic form, which is a type of logic expression including existential quantifiers $\exists$, logic conjunctions $\land$, and disjunctions $\lor$. 
In the logic query, there is a set of anchor entities $V_a$ and values $X_a$. 
These anchor entities and values are the given entities and values in a query, like $Beijing$ in $q_3$ and $California$ in $q_4$ in Figure \ref{fig:teaser}. 
Meanwhile, there are existential quantified entity variables $V_1$, ..., $V_k$, and numerical variables $X_1$,  ..., $X_l$ in a  complex query. 
There is a unique target variable, $V_?$ in each logic query. 
The query intends to find the values of $V_?$, such that there simultaneously exist $V_1, ..., V_k \in \mathcal{V}$ and $X_1, ..., X_l \in \mathcal{N}$ satisfying the logic expression. 
As every logic query can be converted into a disjunctive normal form, 
the complex numerical queries can be expressed in the following formula:

\begin{equation}
\begin{split}
    q[V_?] &= V_?. V_1,...,V_k,X_1,...,X_l :c_1 \lor c_2 \lor ... \lor c_n \\
c_i &= e_{i,1} \land e_{i,2} \land ... \land e_{i,m}.
\end{split}
\end{equation}
Here, $c_i$  are conjunctive expressions of several atomic logic expressions $e_{i,j}$, where each $e_{i,j}$  is any of the following expressions:
$e_{i,j} = r(V,V')$, $e_{i,j} = a(V,X)$, or $e_{i,j} =f(X,X')$,
where the $r$, $a$, and $f$ are the binary expressing entity relation, attribute, and numerical relation respectively. 
They are defined in the previous subsection. 
$V, V', X, X'$ are either anchor entities or values, or existentially quantified variables in $\{V_1, V_2,..., V_k \}$ or $\{X_1, X_2,..., X_l \}$.

\begin{figure}[t]
  \centering
  \includegraphics[width=\linewidth]{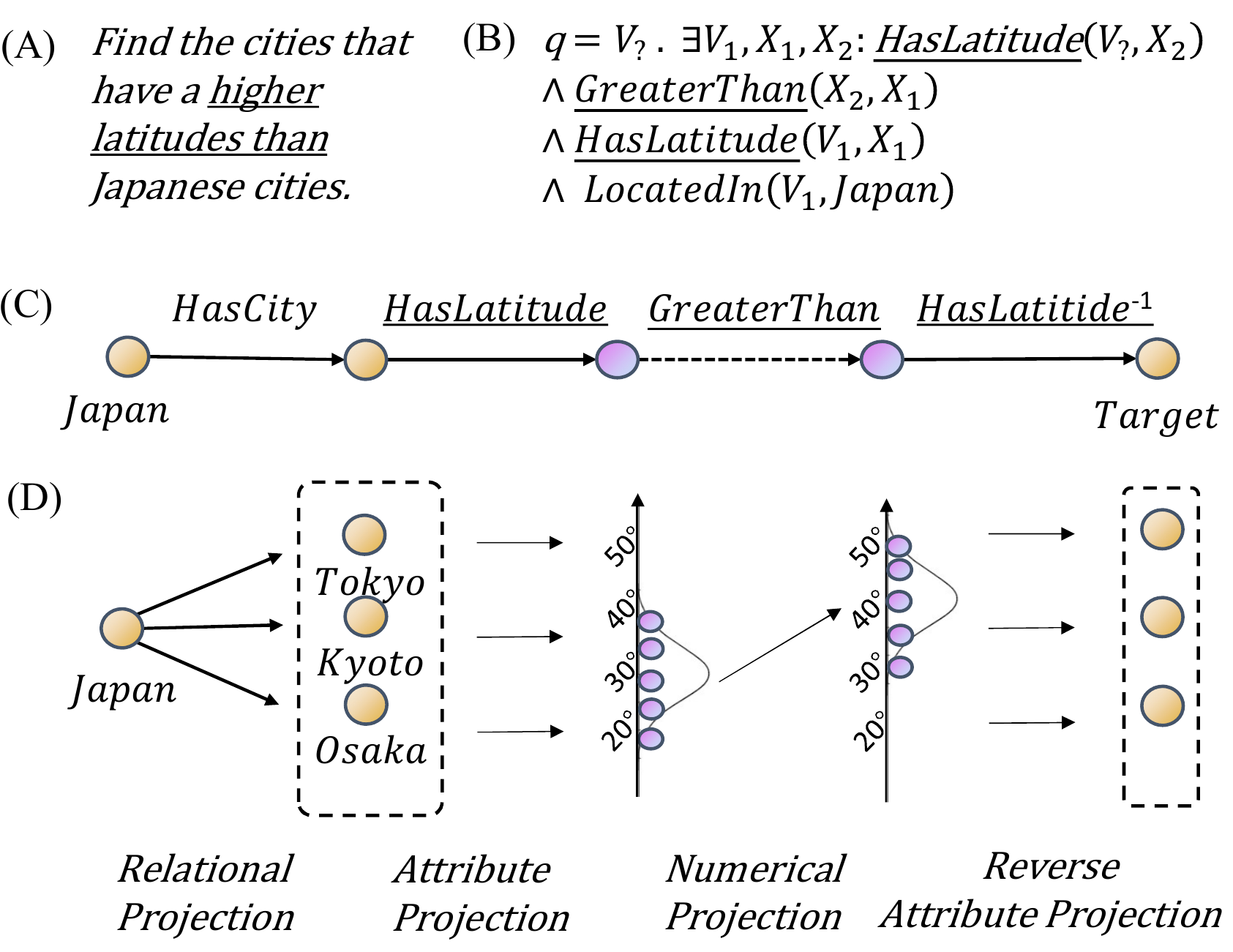}
  \vspace{-0.5cm}
  \caption{An example query demonstrating the Number Reasoning Network (NRN) for complex query with numerical values. 
  (A) The query interpretation; (B) The logic knowledge graph query involving attribute value reasoning; (C) The computational graph parsed from the logic query; (D) The illustrations of four types of projections. The attribute projection, the numerical projection, and the reverse attribute projections are newly introduced for reasoning over values.  
  }
  \label{fig:query_example}
  \vspace{-0.4cm}
 
\end{figure}

\section{Number Reasoning Network for CQA}
\label{section:method}

To deal with the complex queries that involve both entities and numerical values, 
in this section, we propose the number reasoning network (NRN) method. 
Different from the original query encoding methods, in NRN, there are two phases in the  encoding process to encode the entities and numerical values separately. 
One of the phases is the entity encoding phase, where we use the existing query encoding structure, like Box Embedding \cite{ren2020query2box}, to encode intermediate entity set representation.
The other phase is the numerical encoding phase, in which we use parameterized distribution to encode the intermediate numerical representation.  
Here is a sample encoding process of NRN. 
For example, we want to answer the logic query in Figure~\ref{fig:query_example} (B), expressing the meaning of \textit{Find cities that have a higher latitude than Japanese cities. } 
The query will first be parsed into the computational graph as shown in (C). 
To start with, we are given the entity embedding of the anchor entity \textit{Japan}, 
and then a \textit{relational projection} is used to compute the representation of the \textit{cities in Japan}.
In these steps, the encoding process is in the entity phase. 
Then, we will compute the numerical values indicating the  \textit{latitude values of cities in Japan} 
by using \textit{attribute projection}. 
Then we use another encoding structure to encode the set of numerical values in the numerical encoding phase.
Specifically, we use parameterized probabilistic distributions, 
 to estimate the intermediate numerical values. 
When we obtain the intermediate representation of \textit{latitudes of cities in Japan}, we will estimate the distribution of the latitude values that are larger than previous results by using a \textit{numerical projection} of \textit{GreaterThan}.
Finally, a \textit{reverse attribute projection} is computed to find the cities that have such latitude values from the previous step, and the encoding process switches back to the entity phase. 
In the following part of this section, we are going to first give the full definition of the NRN computational graph, 
and then introduce the query encoding structure for both entity and numerical variables. 
Finally, we discuss how to parameterize the operations in the computational graph of NRN.

\subsection{Computational Graph for NRN}

In the query encoding process for NRN, each intermediate state in the encoding process corresponds to either a set of entities or a set of attribute values. 
For the computational graph of entity-only complex queries, only relational projections and logic operations appear. 
Their definitions are as follows:

\begin{itemize}
\item {\textit{Relational Projection}}:  Given a set of entities $S$ and a relation between entities $r\in \mathcal{R}$, the relational projection will produce all the entities that hold a relation r with any of the entities in $S$. Thus the relational projection can be described as
$P_r (S)=\{v \in \mathcal{V} |v' \in S,r(v',v)=1\}$;
\item {\textit{Intersection}}: Given a set of entities or values $S_1,S_2,..., S_n$, this logic operation of intersection computes the intersection of those sets as  $ \cap_{i=1}^n S_i  $;

\item {\textit{Union}}: Given a set of entities or values $S_1  ,S_2  ,...,S_n$, the operation computes the union of   $ \cup_{i=1}^n S_i$   .
\end{itemize}

As shown in Figure \ref{fig:query_example} (D), to compute the computational graph involving numerical attributes, we need to define three types of new projections.
An \textit{attribute projection} is to compute a set of attribute values of a given set of entities. 
A \textit{reverse attribute projection} is to find a set of entities that have the attribute values in a given set of values. 
A \textit{numerical projection} is used to find a set of values that holds certain constraints with another set of values, for example, \textit{GreatThan}, \textit{EqualTo}, etc. 
Their formal definitions are as follows:

\begin{itemize}
\item {\textit{Attribute Projection}}: Given a set of entities S and a numerical attribute type $a$, the attribute projection gives all the values that can be achieved under the attribute type a by any of the entities in $S$. Then the attribute projection can be described by  $P_a(S)=\{x \in  \mathcal{N} | v \in S, a(v,x)=1\}$;

\item {\textit{Reverse Attribute Projection}}: Given a set of values $S \subset  \mathcal{N}$  and a numerical attribute type $a$, the reverse attribute projection gives all the entities whose attribute a has a value in S. As a result, the reverse attribute projection can be described by the following formula $ P_a ^{-1}(S)= \{v \in \mathcal{V} | x \in S, a(v,x)=1\}$;

\item {\textit{Numerical Projection}}: A numerical projection is used to compute a set of values from a set of existing values. This computation depends on the numerical relation function $f$, and it can be described by $P_f(S)=\{x \in \mathcal{N} | x' \in S, f(x',x)=1\}$;

\end{itemize}

\subsection{Representations and Operations in NRN}
\label{subsection:representation}

When a complex query is iteratively encoded through the computation graph, as shown in Figure \ref{fig:query_example} (C), it is either in the entity encoding phase or in the numerical encoding phase. 
Correspondingly, the intermediate state either represents a set of entities in the knowledge graph or a set of attribute values. 

In previous work on query encoding, various methods were proposed to encode entity variables, but they cannot reasonably encode numerical attributes. 
In this paper, we will focus on how to compute and represent numerical values and how to use them to answer complex queries. 
For simplicity, suppose in the step $i$ and the encoding process is entity phase, then the query embedding is denoted as:
$q_i  \in  R^d$.
Otherwise, if the encoding process is in the numerical phase, queries are represented in a more sophisticated way. 
Previous research shows that directly using the numerical value itself as input to a reasoning model is not effective. 
As a result, we employ static numerical encoding methods to project the numerical values from the real space $R$ to the $d$-dimensional real space $R^d$, and this numerical encoding is denoted as $\psi(\cdot)$. 
Meanwhile, we further assume the intermediate numerical answers are drawn from a multi-variant probabilistic distribution in the $d$-dimensional space. 

If the query encoding process is in the numerical encoding phase,  
the query encoding is represented by a family of probability $p_{\theta_i} (x)$ density functions on $R_d$,
where the parameters are denoted as $\theta_i \in R^k$. 
Meanwhile, when we have queries like {\it Find latitudes of the Canadian cities}, we are not only asking for some numerical quantity, but also the types or the units of measure of the quantity, like  \textit{Degrees}, \textit{Meters}, \textit{Dates}, or \textit{Years}.
Suppose the distribution of type $t$ in the numerical space is described by another distribution $\phi_t$.

\subsubsection{Relational Projection and Logic Operations on Entities}
In the number reasoning network, we adopt established query encoding methods as the backbones to conduct relational projections and logic operations for entities. 
In the implementation, we used three different encoding structures to encode the entity phase \cite{ren2020query2box,hamilton2018embedding, bai-etal-2022-query2particles}. 
Despite their differences, their query embeddings can be uniformly flattened to vectors. 
Then, the query encoding in the entity phase is expressed as a $d$-dimensional vector:
$q_i  \in  R^d.$

\subsubsection{Attribute Projection}
\label{subsubsection:ap}
In the attribute projection, suppose we have query embedding of the entity variable at the $i$-th step of $q_i$, and we want to obtain its representation on the attribute values on the attribute type $a$, we can use a neural structure to parameterize the distribution parameters for the $i+1$-th step,
\begin{equation}
\theta_{i+1}  =F_p (  q_i, a),
\end{equation}
where the $F_p$ is a neural structure whose weights will be optimized in the learning process. 
A typical parameterization of $F_p$ is gated transition, which is originally inspired by the recurrent neural network unit~\cite{bai-etal-2022-query2particles}.
Here are the equations for our parameterization:
\begin{equation}
\label{equation:gated_transition}
\begin{split}
&p_i = W^p_pq_i + b_p^p \\
&z_i = \sigma(W^p_ze_a + U^p_zp_i  + b_z^p), \\
&r_i = \sigma(W^p_re_a + U^p_rp_i + b^p_r), \\
&t_i = \phi(W^p_he_a + U_h(r_i \odot p_i ) + b^p_h),\\
&\theta_{i+1} = (1-z_i) \odot p_i + z_i \odot t_i.
\end{split}
\end{equation}
Here, $e_a \in R^k $ is the embedding used to encode the attribute type $a$. Meanwhile, $\sigma$ and $\phi$ are the sigmoid and hyperbolic tangent functions, and  $\odot$ is the Hadamard product.
Also, $W^p_p \in R^{d\cdot k}$ and $W^p_z, W^p_r, W^p_h, U^p_z, U^p_r, U^p_h \in R^{k\cdot k}$ are parameter matrices.

\subsubsection{Reverse Attribute Projection}
\label{subsubsection:rap}
Suppose we want to find entities that have a numerical attribute value from a given set of values. 
Then we compute the query embedding for the resulting entity set by using the reverse attribute projection operation:
\begin{equation}
q_{i+1}  =F_r  (\theta_i  , a),
\end{equation}
where the $F_r$ is a neural structure whose weights will also be optimized in the learning process.
Equation (\ref{equation:gated_transition})'s gated transition function can serve for  reverse attribute projection. While the weights are substituted with a different set of parameters $W^r_p$, $W^r_z$, $W^r_r$, $W^r_h\in R^{k\cdot d}$ and $U^r_z$, $U^r_r$, $U^r_h \in R^{d\cdot d}$.

\begin{table*}[t]
\caption{The statistics of the three knowledge graphs used to construct the benchmark. }
\vspace{-0.3cm}
\label{tab:kg_stats}
\begin{tabular}{ccccccccc}
\toprule
Graphs & Data Split & \#Nodes & \#Rel. & \# Attr. & \#Rel. Edges & \#Attr. Edges  & \#Num. Edges & \#Edges \\
\midrule
\multirow{3}{*}{FB15k} & Training & 25,106 & 1,345 & 15 & 947,540  & 20,248 & 27,020 & 1,015,056 \\
 & Validation & 26,108 & 1,345 & 15 & 1,065,982 &  22,779 & 27,376 & 1,138,916 \\
 & Testing & 27,144 & 1,345 & 15 & 1,184,426  & 25,311 & 27,389 & 1,262,437 \\
 \midrule
\multirow{3}{*}{DB15k} & Training & 31,980 & 279 & 30 & 145,262  & 33,131 & 25,495 & 237,019 \\
 & Validation & 34,191 & 279 & 30 & 161,978 & 37,269  & 25,596 & 262,112 \\
 & Testing & 36,358 & 279 & 30 & 178,394 & 41,411  & 25,680 & 286,896 \\
 \midrule
\multirow{3}{*}{YAGO15k} & Training & 32,112 & 32 & 7 & 196,616  & 21,732 & 26,616 & 266,696 \\
 & Validation & 33,078 & 32 & 7 & 221,194 & 22,748  & 26,627 & 293,317 \\
 & Testing & 33,610 & 32 & 7 & 245,772  & 23,520 & 26,631 & 319,443 \\
\bottomrule
\end{tabular}
\vspace{-0.3cm}
\end{table*}

\subsubsection{Numerical Projection}
\label{subsubsection:np}
The numerical projections are directly applied to the parameters of distributions. As the numerical constraint of $f$ could be an arbitrary binary function from $R^2$ to $\{0,1\}$, for now, we use the function $F_f$   to express the projection operation:
\begin{equation}
\theta_{i+1}  =F_f  (\theta_i).
\end{equation}
In a typical parameterization, we can use an embedding $e_f$ to encode the characteristics of the numerical function $f$. 
The transition of distribution caused the function $f$ to be learned in the training process. 
For simplicity, we can still use gated transition in Equation (\ref{equation:gated_transition}) while replacing the embedding $e_r$ with $e_f$, and replacing their weights with $W^f_j, U^f_j\in R^{k\cdot k}$, where $j \in \{ z, r, h \}$ respectively.
\subsubsection{Intersection and Union on Attribute Values}
\label{subsubsection:intersection_union}

Suppose we have n attribute variables  $\theta_i^1, \theta_i^2,..., \theta_i^n$  as inputs for intersection operations. Their intersection and union are expressed as:

\begin{equation}
\begin{split}
\theta_{i+1}  =F_i  (\theta_i^1  ,\theta_i^2  ,...,\theta_i^n  ), \\
\theta_{i+1}  =F_u  (\theta_i^1  ,\theta_i^2  ,...,\theta_i^n  ).
\end{split}
\end{equation}
For functions $F_i$ and $F_u$, we use  \texttt{DeepSet} \cite{zaheer2017deep} functions to parameterize them respectively. 
\begin{align}
    \theta_{i+1}  = \texttt{DeepSet} ([\theta_i^1  ,\theta_i^2  ,...,\theta_i^n]).
\end{align}
The intersection and union operations are thus invariant to input variable permutations. 
In our implementation, we use a self-attention network followed by a feed-forward network to implement such permutation invariant \texttt{DeepSet} function. 
Suppose $\theta_i =  [\theta_i^1  ,\theta_i^2  ,...,\theta_i^n]$, then 
their formulas are as follows:

\begin{align}
a_{i+1} = & \texttt{Attn}(W _q \theta^T_i , W _k \theta^T_i, W _v \theta^T_i )^T\\
\theta_{i+1}= &\texttt{MLP}(a_i).
\end{align}
The $W _q, W _k, W _v\in R^{k\times k}$ are parameters used for modeling the input Query, Key, and Value for the attention module \texttt{Attn}.
The \texttt{Attn} represents the scaled dot-product attention,
\begin{align}
    \texttt{Attn}(Q, K, V) = \texttt{Softmax}(\frac{QK^T}{\sqrt{d}}) V.
\end{align}
Here, the $Q$, $K$, and $V$ represent the input Query, Key, and Value for this attention layer. The \texttt{MLP} here denotes a multi-layer perceptron layer with \texttt{ReLU} activation.


\subsubsection{Learning Attribute Reasoning}

As there are two types of queries with different types of target variables, we define two loss functions respectively. Suppose there are total $I$ steps in the computational graphs, the target variable is an entity variable, and its query embedding is $q_I$, then we compute the normalized probability of the entity $v$ being the correct answer to the query $q$ by using the \texttt{softmax} function on all similarity scores,

\begin{equation}
p(q,v)=  \frac{e^{<q_I, e_v>}}{  \sum _{v' \in N} e^{ < q_I , e_{v'}>}  }   .  
\end{equation}

Then the objective function of entities is expressed as
\begin{equation}
L_{E}  =  - \frac{1}{N} \sum_{j=1}^N    \log p(q^{(j)}, v^{(j)}).
\end{equation}
Here, each $(q^{(i)}, v^{(i)})$ is one of the positive query-answer parties, and there are $N$ such pairs.

Meanwhile, for the numerical variables, we are going to use the maximize a posteriori probability (MAP) estimation to derive an objective function for type-aware attribute reasoning. 
Suppose the parameters of the attribute distribution of the final step are $\theta_I$, the positive answer is $v$, and the answer has a value type $t$. 
The $\theta_I$ is computed from $\theta_i$ and $q_i$ where $i \in \{1,2,3,..., I-1\}$ by using a series of projections like $F_p$ ,$F_r$, $F_f$, and logic operations $F_i$ and $F_u$.
Then our goal is to optimize $\theta_I$ given $v$ and $t$. Then suppose $f$ and $g$ are the conditional distribution of $v$ given  $\theta_I$ and $t$, and $\theta_I$ given $t$, and meanwhile $v$ is conditional independent of $t$ given $\theta_I$.
Then according to MAP estimation,  
\begin{equation}
\begin{split}
\hat {\theta_I} (v, t) &= \argmax_{\theta_I} f(\theta_I | v, t)\\
&= \argmax_{\theta_I} \frac{f(v|\theta_I, t)g(\theta_I|t)}{\int_{\theta_I}f(x|\theta, t)g(\theta|t)d\theta} \\
&= \argmax_{\theta_I} f(v|\theta_I, t)g(\theta_I|t)\\
&= \argmin_{\theta_I} (-\log f(v|\theta_I) - \log g(\theta_I|t)).
\end{split}
\end{equation}

Suppose $(q^{(j)}, v^{(j)})$ is a pair of positive query-answer pairs, and we have totally M such pairs. Suppose the value type of $v^{(i)}$ is $t^{(i)}$. 
Using the previous defined notions, then  $f(v^{(j)}|\theta_I^{(j)})$ can be written as $p_{\theta_I^{(j)}}(v^{(j)})$, and $g(\theta_I|t))$ can be written as $\phi_{t^{(j)}}(\theta_I^{(j)})$.
Then, the loss of explicit queries can be described by
\begin{equation}
L_A  =  \frac{1}{M}  \sum_{j=1}^M (-\log p_{\theta_I^{(j)}}(\psi(v^{(j)}))  -\log \phi_{t^{(j)}}(\theta_I^{(j)})),
\label{equation:attribute_loss}
\end{equation}
where the $\psi(\cdot)$ is numerical encoding defined on the domain of $R$ with the range of $R^d$.
In the training process, we alternatively optimize these two loss functions to train all parameters in different neural structures together.

\begin{table*}[t]
\caption{The statistics of the queries sampled from the FB15k, DB15k, and YAGO15k knowledge graphs.  }
  \vspace{-0.3cm}
\label{tab:queries_stats}
\begin{tabular}{ccccccccccc}
\toprule
Graphs & Data Split & 1p & 2p & 2i & 3i & pi & ip & 2u & up & All \\
\midrule
\multirow{3}{*}{FB15k} & Training & 304,633 & 138,192 & 226,729 & 288,874 & 260,057 & 233,834 & 284,301 & 284,931 & 2,021,551 \\
 & Validation & 8,271 & 15,860 & 23,359 & 28,836 & 25,081 & 22,930 & 29,187 & 29,210 & 182,734 \\
 & Testing & 7,969 & 15,431 & 23,346 & 28,865 & 24,810 & 22,232 & 29,212 & 29,274 & 181,139 \\
 \midrule
\multirow{3}{*}{DB15k} & Training & 124,851 & 99,698 & 140,427 & 190,413 & 171,353 & 163,687 & 190,364 & 194,244 & 1,275,037 \\
 & Validation & 3,529 & 10,388 & 9,792 & 13,817 & 14,594 & 16,651 & 19,512 & 19,792 & 108,075 \\
 & Testing & 3,387 & 10,047 & 9,914 & 14,603 & 14,642 & 15,897 & 19,504 & 19,773 & 107,767 \\
 \midrule
\multirow{3}{*}{YAGO15k} & Training & 84,014 & 76,238 & 136,282 & 183,850 & 162,712 & 145,994 & 183,963 & 183,459 & 1,156,512 \\
 & Validation & 2,833 & 7,986 & 10,757 & 16,884 & 13,485 & 13,899 & 18,444 & 19,105 & 103,393 \\
 & Testing & 2,713 & 7,949 & 10,935 & 17,171 & 13,481 & 13,526 & 18,433 & 18,997 & 103,205 \\
 \bottomrule
\end{tabular}
\end{table*}

\section{Benchmark Construction}
\label{section:benchmark}
In this section, we construct three benchmark datasets for evaluating the performance of numerical complex query answering.

\subsection{Knowledge Graphs}

We use FB15k, DB15k, and YAGO15k \cite{kotnis2018learning} as the knowledge graphs to create our benchmarks.
These KGs are publicly available, and they include both triples describing entity relations and numerical values.
For each knowledge graph, we first randomly divide the edges into training, validation, and testing edges with a ratio of 8:1:1 respectively. 
Then, the training graph $\mathcal{G}_{train}$, validation graph $\mathcal{G}_{val}$, and testing graph $\mathcal{G}_{test}$ are aggregated from the training edges, training+validation edges, and training+validation+testing edges respectively. 
The detailed statistics of the knowledge graphs are shown in Table \ref{tab:kg_stats}. 
The columns \#Rel. and \#Attr. denote the number of entity relation types and the number of attribute types respectively. \#Rel. Edges, \#Attr. Edges,  and  \#Num. Edges represent the total number of edges describing entity relations, attribute values, and numerical relations between  values.

\begin{figure}[t]
  \centering
  \includegraphics[width=0.8\linewidth]{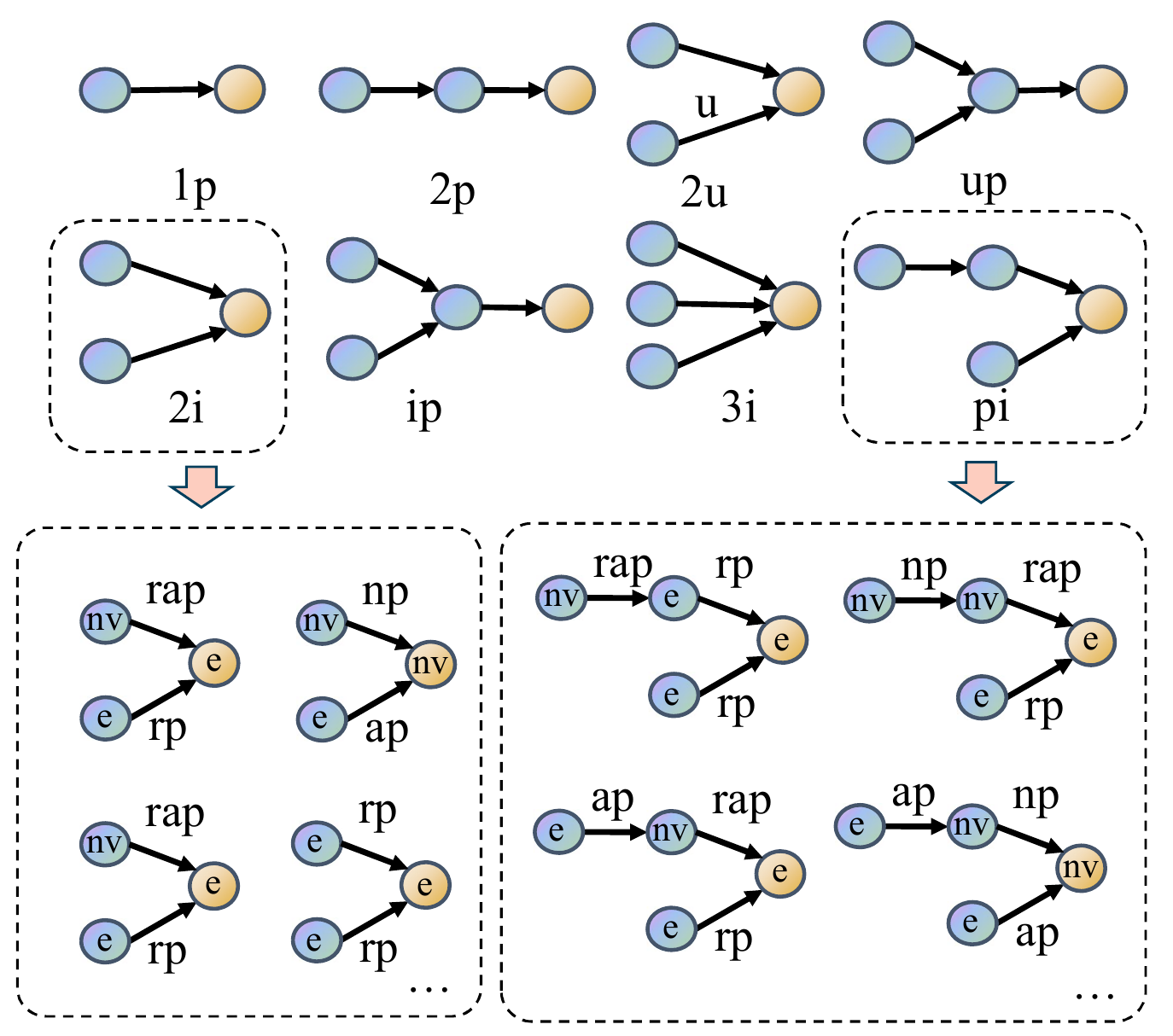}
    \vspace{-0.2cm}
  \caption{Eight general query types. Each general query type corresponds to multiple specific query types.}
  \label{fig:query_types}
  \vspace{-0.4cm}
\end{figure}

\begin{figure}[t]
  \centering
  \includegraphics[width=0.95\linewidth]{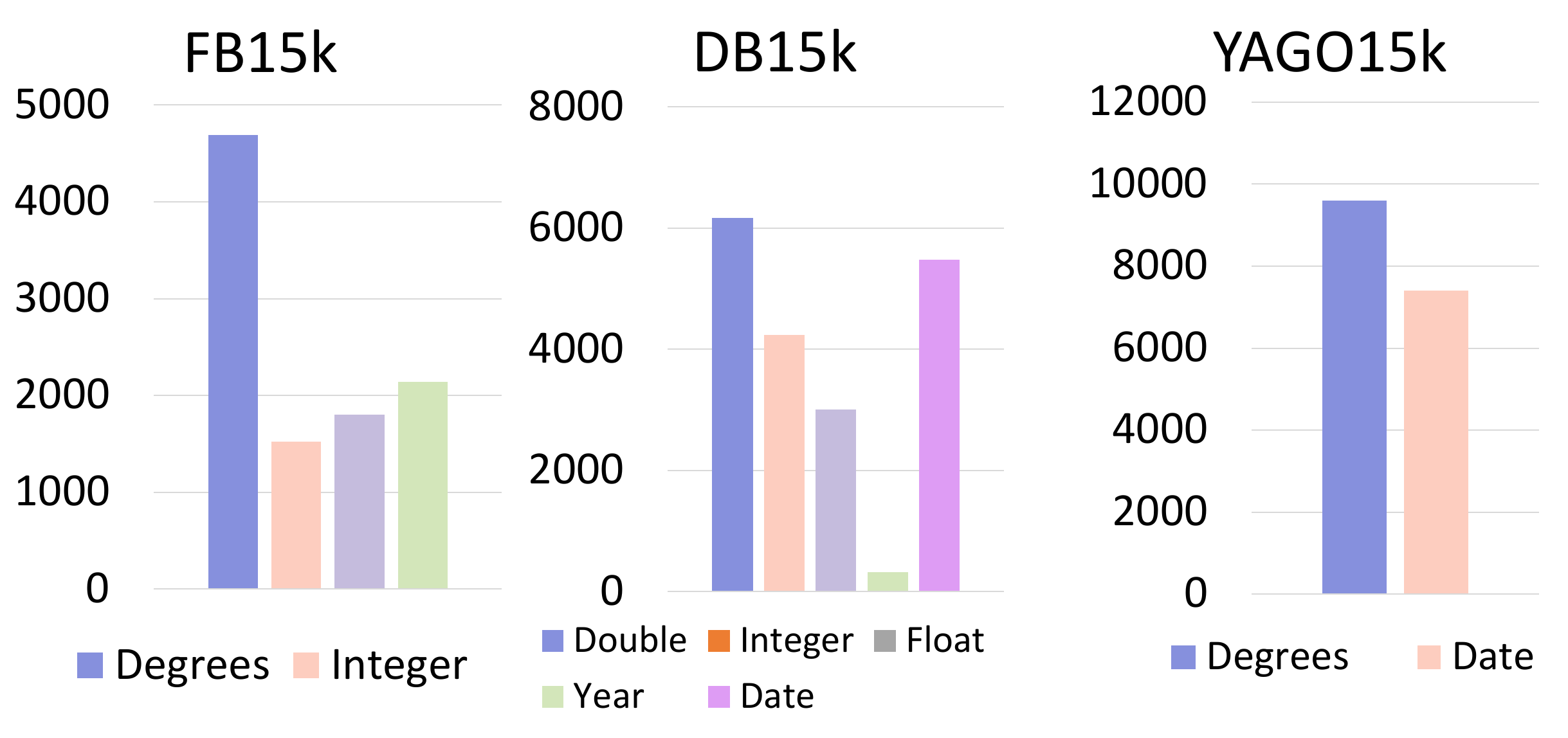}
  \vspace{-0.3cm}
  \caption{The distribution of the types of numerical values on three different knowledge graphs.}
   \vspace{-0.4cm}
   \label{fig:value_types}
\end{figure}

\subsection{Query Types}
Following previous work \cite{hamilton2018embedding}, we used the following eight types of query as general query types. For each general query type, each edge represents a projection, and each node either represents either a set of entities or numerical values.
Based on these query types and knowledge graph structures, we further derive various specific sub-types. 
For example, the query type \textit{pi} can be instantiated into various types of specific types of queries. 
As shown in Figure \ref{fig:query_types}, \textit{e} stands for an entity, \textit{nv} stands for numerical values, and \textit{ap}, \textit{rap}, \textit{rp}, and \textit{np} are the abbreviations of \textit{attribute projection}, \textit{reverse attribute projection}, \textit{relational projection}, and \textit{numerical projections} respectively. 
The general query types are shown in Table~\ref{tab:query_types}.

\subsection{Sampling of Complex Attribute Queries}

In this section, we show how to sample numerical complex queries. 
Before sampling, we added dummy numerical edges to the graphs to evaluate the performance on numerical relations. 
For simplicity, we consider seven types of numerical relations: \textit{EqualTo}, \textit{SmallerThan}, \textit{GreaterThan}, \textit{TwiceEqualTo}, \textit{ThreeTimesEqualTo}, \textit{TwiceGreaterThan}, and \textit{ThreeTimesGreaterThan}. 
These numerical relations are randomly added to the attribute value pairs with the same value type and satisfy the corresponding constraints. 
For each type of numerical relation, we set an upper limit of four thousand edges on each graph. 
The total number of numerical edges added is shown in Table \ref{tab:kg_stats}.
There are two general steps in sampling the complex attribute queries. 
In the first step, we are given eight types of general queries with abbreviations of \textit{1p},\textit{2p},\textit{2i},\textit{3i},\textit{ip},\textit{ip},\textit{2u}, and \textit{up}.
Their definitions are given in Table \ref{fig:query_types}.
Given the general query type and a knowledge graph $G$, a sampling algorithm is used to generate a complex attribute query that has at least one answer. 
The details of the algorithm are given in the Appendix. 
In the second step, given a complex attribute query from the first step, we conduct a graph search and find its corresponding answers on the KG.

\begin{table}[t]
\caption{The abbreviations of the  query types and structures. }
\label{tab:query_types}
\small
\begin{tabular}{c|l|l}
\toprule
Abbr. & General Query Type & Specific Query Type Example \\ 
\midrule
1p & (p,(e)) & (rap,(nv)) \\
2p & (p,(p,(e))) & (rap, (np,(nv))) \\
2i & (i,(p,(e)),(p,(e))) & (i, (np,(nv)), (ap, (e))) \\
3i & (i,(p,(e)),(p,(e)),(p,(e))) & (i, (np,(nv)), (ap, (e)), (ap, (e))) \\
ip & (p,(i,(p,(e)),(p,(e)))) & (rp, (i, (np,(nv)), (ap, (e)))) \\
pi & (i,(p,(e)),(p,(p,(e)))) & (i, (np,(nv)), (ap, (rp, (e)))) \\
2u & (u,(p,(e)),(p,(e))) & (u, (np,(nv)), (ap, (e))) \\
up & (p,(u,(p,(e)),(p,(e)))) & (rp, (u, (np,(nv)), (ap, (e))))\\
\bottomrule
\end{tabular}
\vspace{-0.4cm}
\end{table}

For each knowledge graph, we have constructed their corresponding Training, Validation, and Testing graphs. 
Then, the first step of the previously described query sampling process is conducted on these three data splits independently. 
After this, we obtained the training queries, validation queries, and test queries respectively. 
Then, For the training queries, we conduct a graph search on the training graph to find corresponding training answers.
For the validation queries, we search their answers on both the training graph and the validation graph, and we only use the queries that have a different number of training and validation answers.
Finally, for the testing queries, the graph searches are conducted on training, validation, and testing respectively. 
Similarly, we only use the testing queries that have a different number of answers on validation and test graphs.
After these two steps are done, the statistics of the resulting training, validation, and testing queries of three knowledge graphs are shown in Table \ref{tab:queries_stats}.

\section{Experiment}
\label{section:experiment}
In this section, we implement the proposed number reasoning network (NRN) and compare it to existing query encoding models.

\subsection{Baseline Methods}
\label{section:querty_encoding_methods}
NRN uses previous query encoding methods to encode entities during the entity encoding phase.
It can be combined with a wide range of entity encoding structures that can be transformed into a fixed length of real-valued vectors. 
We use the following methods:

\begin{itemize}
    \item GQE \cite{hamilton2018embedding}: The graph query encoding  model encodes a complex logic query into a single vector;
    \item Q2B \cite{ren2020query2box}: The Query2Box model encodes a complex query into hyper-rectangles in the embedding space; 
    \item Q2P \cite{bai-etal-2022-query2particles}: The Query2Particles model encodes a complex logic query into multiple vectors. 
\end{itemize}

\subsection{Numerical Encoding}
As explained in \S \ref{subsection:representation}, we use numerical encoding methods to project the attribute values from $R$ to $R^D$ to obtain number embedding for NRN. 
We use the following two deterministic numerical encodings.

\begin{itemize}
    \item DICE \cite{sundararaman-etal-2020-methods}: DICE is a deterministic independent-of-corpus word embedding of numbers. 
    DICE managed to use a linear mapping from the numerical values $v$ to an angle $\alpha \in [0, \pi]$, and then conduct a polar-to-Cartesian transformation in the $D$-dimensional space according to the following formula:
    \begin{equation}
        \psi(v)_d = \left\{ \begin{array}{lcr}
          \sin^{d-1}(\alpha) \cos(\alpha), & 1\leq d<D\\
           \sin^{D}(\alpha), & d= D\\
        \end{array} \right.
    \end{equation}
    \item  Sinusoidal \cite{vaswani2017attention}: The Sinusoidal encoding of value is first proposed in encoding token position in the transformer model. In this paper, we found that this function is also effective in encoding numerical values. Its formula is expressed by:
    \begin{equation}
        \psi(v)_d = \left\{ \begin{array}{lrr}
          \sin(\frac{v}{n^{d/D}}), &  d \equiv  0  & \mod 2\\
           \cos(\frac{v}{n^{(d-1)/D}}), & d  \equiv 1  &  \mod  2\\
        \end{array} \right.
    \end{equation}
\end{itemize}
These two functions are both deterministic and preserve geometric properties from the original value space to the embedding space.

\begin{table}[t]
\footnotesize
\caption{The main experiment results of numerical complex query answering. NRN constantly outperforms the baseline methods on Hit@1, Hit@3, Hit@10, and MRR. }
\vspace{-0.2cm}
\label{tab:general_performance}
\begin{tabular}{c|c|l|rrrr}
\toprule
\multicolumn{1}{l|}{Dataset} & \multicolumn{1}{l|}{Entity} & Value  & H@1 & H@3 & H@10 & MRR \\ 
\midrule
\multirow{9}{*}{FB15k} & \multirow{3}{*}{GQE} & Baseline & 13.56 & 23.47 & 34.80 & 20.80 \\ 
 &  & NRN + DICE & 14.16 & 24.69 & 36.18 & 21.72 \\
  &  & NRN + Sinusoidal & \textbf{14.33} & \textbf{24.81} & \textbf{36.25} & \textbf{21.86} \\   \cmidrule{2-7} 
 & \multirow{3}{*}{Q2B} & Baseline & 14.49& 25.96&38.20&22.56 \\ 
 &  & NRN    + DICE & 15.55&	27.31&	39.24&	23.72 \\
  &  & NRN + Sinusoidal & \textbf{15.78}&\textbf{27.65}&\textbf{39.54}&\textbf{23.96} \\   \cmidrule{2-7} 
 & \multirow{3}{*}{Q2P} & Baseline & 15.54 & 25.50 & 36.88 & 22.84 \\ 
 &  & NRN    + DICE & 16.98 & 27.02 & 38.36 & 24.27 \\
  &  & NRN + Sinusoidal & \textbf{17.72} & \textbf{27.96} & \textbf{39.23} & \textbf{25.09} \\  \cmidrule{1-7}
\multirow{9}{*}{DB15k} & \multirow{3}{*}{GQE} & Baseline & 6.86 & 12.98 & 21.28 & 11.73 \\ 
 &  & NRN    + DICE & 7.26 & 13.36 & 21.81 & 12.15 \\
  &  & NRN +  Sinusoidal & 7.30 & 13.63 & \textbf{22.32} & \textbf{12.34} \\ 
   \cmidrule{2-7}
   & \multirow{3}{*}{Q2B} & Baseline &7.47&14.17&22.81&12.66 \\ 
 &  & NRN  + DICE & \textbf{8.07}&\textbf{15.21}&\textbf{24.22}&\textbf{13.53} \\
  &  & NRN + Sinusoidal & 8.05&15.04&24.05&13.45 \\   \cmidrule{2-7} 
 & \multirow{3}{*}{Q2P} & Baseline & 8.41 & 14.57 & 22.84 & 13.31 \\ 
 &  & NRN    + DICE & 8.81 & 15.10 & 23.27 & 13.73 \\
  &  & NRN +  Sinusoidal & \textbf{9.13} & \textbf{15.69} & \textbf{24.22} & \textbf{14.25} \\ \cmidrule{1-7}
\multirow{9}{*}{YAGO15k} & \multirow{3}{*}{GQE} & Baseline & 10.56 & 18.12 & 27.65 & 16.35 \\ 
 &  & NRN    + DICE & 11.68 & 19.49 & 29.05 & 17.59 \\
  &  & NRN + Sinusoidal & \textbf{11.78} & \textbf{19.74} & \textbf{29.12} & \textbf{17.73} \\ \cmidrule{2-7}
  
  & \multirow{3}{*}{Q2B} & Baseline &13.47&22.67&32.56&20.02 \\ 
 &  & NRN    + DICE & 13.94&23.74&33.56&20.77 \\
  &  & NRN + Sinusoidal & \textbf{14.41}&\textbf{23.96}&\textbf{33.78}&\textbf{21.13} \\   \cmidrule{2-7} 
 & \multirow{3}{*}{Q2P} & Baseline & 6.70 & 11.98 & 20.11 & 11.29 \\ 
 &  & NRN    + DICE & 9.80 & \textbf{16.98 }& \textbf{26.74} & \textbf{15.53} \\
  &  & NRN + Sinusoidal &\textbf{ 9.90 }& 16.83 & 26.43 & 15.49 \\ 
\bottomrule
\end{tabular}
\vspace{-0.4cm}
\end{table}

\begin{table*}[t]

\caption{The mean reciprocal ranking (MRR) results for different types of numerical complex queries. }
\vspace{-0.3cm}
\small
\label{tab:detailed_performance}
\begin{tabular}{c|c|l|rrrrrrrr}
\toprule
Dataset & Entity Encoding & Values Encoding & 1p & 2p & 2i & 3i & pi & ip & 2u & up \\ \midrule
\multirow{9}{*}{FB15k} & \multirow{3}{*}{GQE} & Baseline & 30.63 & 7.54 & 32.70 & 39.61 & 21.57 & 8.87 & 7.18 & 4.17 \\  
 &  & NRN   + DICE & \textbf{31.75} & \textbf{8.16} & 33.96 & 40.32 & 22.70 & 9.80 & \textbf{8.45} & \textbf{4.66} \\
  &  & NRN + Sinusoidal & 31.69 & 8.15 & \textbf{33.99} & \textbf{41.19} & \textbf{23.10} & \textbf{10.10} & 7.57 & 4.51 \\
  \cmidrule{2-11} 
 & \multirow{3}{*}{Q2B} & Baseline & 35.16&7.12&35.39&41.31&22.99&9.17&12.76&4.02\\  
 &  & NRN   + DICE &36.96&7.66&36.50&42.30&24.17&\textbf{10.41}&14.53&\textbf{4.79}\\
  &  & NRN + Sinusoidal & \textbf{37.01}&\textbf{7.74}&\textbf{36.88}&\textbf{42.72}&\textbf{24.25}&10.09&\textbf{15.61}&4.54 \\
  \cmidrule{2-11} 
 & \multirow{3}{*}{Q2P} & Baseline & 39.29 & 12.45 & 30.40 & 35.29 & 20.09 & 11.05 & 21.25 & 7.80 \\  
 &  & NRN   + DICE & 41.45 & 12.65 & 32.91 & 37.61 & 22.02 & 12.85 & 21.74 & \textbf{8.03} \\
  &  & +  Sinusoidal & \textbf{42.75} & \textbf{12.87} & \textbf{33.71} & \textbf{38.75} & \textbf{23.14} & \textbf{13.23} & \textbf{23.16} & 7.89 \\ 
\hline
\multirow{9}{*}{DB15k} & \multirow{3}{*}{GQE} & Baseline & 9.83 & 2.41 & 18.83 & 34.48 & 11.21 & 2.11 & 1.86 & 1.94 \\  
 &  & NRN   + DICE & \textbf{10.46} & \textbf{2.58} & \textbf{20.17} & 34.80 & 11.88 & 2.48 & 1.94 & 2.03 \\
 &  & NRN + Sinusoidal & 10.29 & 2.53 & 20.14 & \textbf{35.46} & \textbf{12.50} & \textbf{2.52} & \textbf{2.08} & \textbf{2.14} \\
   \cmidrule{2-11} 
   & \multirow{3}{*}{Q2B} & Baseline & 10.18&2.53&20.81&36.29&12.57&2.72&2.06&1.67 \\  
 &  & NRN   + DICE & \textbf{11.01}&2.66&22.38&\textbf{37.47}&\textbf{14.17}&\textbf{3.09}&\textbf{2.62}&\textbf{2.20} \\
  &  & NRN + Sinusoidal & 10.96&\textbf{2.71}&\textbf{22.60}&37.44&13.81&3.05&2.41&2.13 \\
  \cmidrule{2-11} 
 & \multirow{3}{*}{Q2P} & Baseline & 14.44 & \textbf{3.96} & 20.67 & 33.85 & 13.64 & \textbf{3.48} & 4.66 & \textbf{2.96} \\  
 &  & NRN   + DICE & 14.58 & 3.81 & 22.15 & 35.36 & 13.90 & 2.83 & \textbf{4.94} & 2.67 \\
  &  & NRN + Sinusoidal & \textbf{14.71} & 3.81 & \textbf{23.75} & \textbf{36.66} & \textbf{14.47} & 2.96 & 4.63 & 2.81 \\
\midrule
\multirow{9}{*}{YAGO15k} & \multirow{3}{*}{GQE} & Baseline & 14.28 & 3.18 & 32.83 & 37.83 & 15.91 & 4.89 & 3.62 & 1.82 \\  
 &  & NRN   + DICE & \textbf{15.21} & 4.20 & 35.08 & \textbf{40.32} & 17.01 & \textbf{5.71} & 3.95 & 1.95 \\
  &  & NRN + Sinusoidal & 14.79 &\textbf{ 4.23} & \textbf{35.68} & 39.64 & \textbf{18.29} & 5.65 & 4.23 & \textbf{1.96} \\ 
 \cmidrule{2-11} 
 & \multirow{3}{*}{Q2B} & Baseline & 18.84&3.91&38.62&44.67&18.72&7.31&6.90&2.47\\  
 &  & NRN   + DICE & 20.83&4.53&38.87&\textbf{45.19}&\textbf{20.61}&7.69&8.72&2.73 \\
  &  & NRN + Sinusoidal & \textbf{21.40}&\textbf{4.59}&\textbf{39.72}&45.16&19.62&\textbf{7.90}&\textbf{9.05}&\textbf{2.82} \\ 
  \cmidrule{2-11} 
 & \multirow{3}{*}{Q2P} & Baseline & 19.84 & 4.64 & 17.42 & 19.34 & 11.14 & 4.33 & 9.92 & 3.07 \\  
  &  & NRN   + DICE & 22.68&5.36&25.45&\textbf{29.70}&14.07&\textbf{5.67}&11.73&3.47 \\
  &  & NRN + Sinusoidal & \textbf{22.97}&\textbf{5.70}&\textbf{25.70}&29.04&\textbf{14.38}&5.40&\textbf{11.87}&\textbf{3.51} \\
 
\bottomrule
\end{tabular}
\vspace{-0.3cm}
\end{table*}

\subsection{Evaluation Metrics}

For a testing query $q$, the training, validation, and testing answers are denoted as $[q]_{train}$, $[q]_{val}$, and $[q]_{test}$ respectively. 
In the experiment, we evaluate the generalization capability of the attribute reasoning models by computing the rankings of the answers that cannot be directly searched from an observed knowledge graph. 
Suppose $[q]_{val}/[q]_{train}$ represents the answer set of the query $q$ that can be searched from the validation graph but cannot be searched from the training graph. 
Similarly, $[q]_{test}/[q]_{val}$ is the set of answers that can be found on the test graph but cannot be found in the validation graph. 
Then, the evaluation metrics of the test query $q$ can be expressed in the following equation:

\begin{equation}
   \text{Metric}(q) = \frac{1}{|[q]_{test}/[q]_{val}|} \sum_{v \in [q]_{test}/[q]_{val}} m(rank(v)).  \\ \label{equa:metrics}
\end{equation}
When the evaluation metric is Hit@K, the $m(r)$ in Equation (\ref{equa:metrics}) is defined as $m(r) = \textbf{1}[r\leq K]$. In other words, $m(r)=1$ if $r\leq K $, otherwise $m(r) =0$. Meanwhile, if the evaluation metric is mean reciprocal ranking (MRR), then the $m(r)$ is defined as $m(r) = \frac{1}{r}$.
Similarly, if $q$ is a validation query, its generalization capability is evaluated on $[q]_{val}/[q]_{train}$ instead of $[q]_{test}/[q]_{val}$. 
In our experiment, we train all the models by using the training queries, and we select the hyper-parameters by using the validation queries. The evaluation is then finally conducted on the testing queries. 

\begin{table*}[t]
\caption{The average training time  and inference time per query in milliseconds (ms) of complex numerical queries. Although the training and inference are slower than the baseline, NRN still achieves high training and inference speed. }
\vspace{-0.3cm}
\label{tab:computing_time}
\begin{tabular}{@{}c|cc|cc|cc|cc@{}}
\toprule
\multirow{2}{*}{Time (ms)} & \multicolumn{2}{c|}{FB15k} & \multicolumn{2}{c|}{DB15k} & \multicolumn{2}{c|}{YAGO15k} & \multicolumn{2}{c}{Average} \\
 & Training & Inference & Training & Inference & Training & Inference & Training & Inference \\ \midrule
Baseline & 0.090 & 0.042 & 0.078 & 0.031 & 0.088 & 0.040 & 0.085 & 0.037 \\
NRN & 0.253 & 0.149 & 0.229 & 0.134 & 0.255 & 0.152 & 0.245 & 0.145 \\ \bottomrule
\end{tabular}
\vspace{-0.3cm}
\end{table*}

\vspace{-0.1cm}
\subsection{Experiment Results}

As discussed in \S\ref{section:querty_encoding_methods}, we use three different query encoding methods GQE~\cite{hamilton2018embedding}, Q2B~\cite{ren2020query2box}, and Q2P~\cite{bai-etal-2022-query2particles} 
as the backbone for entities in our framework. 
Meanwhile, to evaluate the performance of  NRN, we also use the original version of the GQE, Q2B, and Q2P methods as baselines. 
In these baselines, all attribute values in the knowledge graphs are treated in the same way as other entities in the knowledge graph. 
In our proposed NRN framework, we use two ways of incorporating attribute value information in the reasoning model. 
First, we use the DICE \cite{sundararaman-etal-2020-methods} and Sinusoidal \cite{vaswani2017attention} function to encode attribute values as the numerical encoding function $\psi(v)$ in Equation (\ref{equation:attribute_loss}). 
Meanwhile, for simplicity, we choose to use the multi-variant Gaussian distributions as the distribution family used for the distribution of numerical queries $p_{\theta}(x)$ and the distribution of the parameters given the value type $\phi_{t}(\theta)$ in Equation (\ref{equation:attribute_loss}). 
Experiment results are reported in Table \ref{tab:general_performance} and Table \ref{tab:detailed_performance}.

Table \ref{tab:general_performance} reports the averaged  results in Hit@1, Hit@3, Hit@10, and MRR. 
The experiment results show that the number reasoning networks constantly outperform their baseline reasoning methods on three different knowledge graphs under four different evaluation metrics. 
The difference between the baseline query encoding method and our number reasoning network is that we use parameterized distributions to encode the numerical attribute values instead of using a unified query encoding structure. 
As a result, from the empirical results, we can conclude that parameterized distributions are effective encoding structures for numerical attributes. 
On the other hand, the Sinusoidal function performs on par or even better than DICE on different KGs under various metrics. 
Thus Sinusoidal is a simple yet effective method to encode the magnitude of knowledge graph attribute values.
Moreover, the mean reciprocal rank (MRR) scores for different types of queries are shown in Table \ref{tab:detailed_performance}.  
Empirical results show that NRN performs better than the baselines on almost every type of query on the three datasets. 

\vspace{-0.1cm}
\subsection{Computing Time Analysis}
Further analysis on computing time is also conducted. 
In this experiment, we measure the average training and inference time on FB15k, DB15k, and YAGO15K with the Q2P. The experiment results are shown in Table \ref{tab:computing_time}. 
We run our experiments on the RTX2080ti graphics card with a batch size of 1024. Then we compute the average training and inference time per query using the batch processing time divided by the batch size. 
The empirical results show that the average inference time of NRN is 0.11 milliseconds longer than the baseline, and the training time is 0.16 milliseconds longer. 
Although the computing time is longer than the baseline model, the NRN is still fast and effective.

\section{Related Work}

The query encoding methods are closely related to this work. 
Some methods use different geometric to encode queries. 
The GQE \cite{hamilton2018embedding} model encodes a KG query as a vector in embedding space. 
Query2Box \cite{ren2020query2box} encodes a KG query as a hyper-rectangle in an embedding space. It can be used for answering existential positive first-order (EPFO) complex queries.
EmQL \cite{sun2020faithful} proposes to use count sketches to improve the faithfulness of the query embeddings by using count-min sketches. 
ConE \cite{zhang2021cone} improves the Q2B by using cone embeddings. The cone embeddings are able to express negation and encode arbitrary first-order logic queries.
CylE \cite{DBLP:conf/eacl/NguyenFLS23} proposes to use cylinders to encode complex queries. 
Q2P \cite{bai-etal-2022-query2particles} encodes a logic query into multiple vectors in the embeddings space. 
Meanwhile, HypE \cite{choudhary2021self} encodes the complex queries as hyperbolic query embedding. 
\citet{DBLP:journals/corr/abs-2305-04034} propose Wasserstein-Fisher-Rao Embedding for CQA.
Meanwhile, some methods try to encode logic queries into specific probabilistic structures. 
Beta Embedding \cite{ren2020beta} proposes to use multiple Beta distributions to encode logic knowledge graph queries expressed in First-order logic queries. 
\citet{DBLP:conf/emnlp/YangQLLL22} propose to use Gamma distributions to encode complex queries. 
Then, PERM \cite{choudhary2021probabilistic} embeddings encode the complex queries by using a mixture of Gaussians. 
Recently, Line Embedding \cite{huang2022line} is proposed to relax the distributional assumptions by using a logic space transformation layer to convert probabilistic embeddings to LinE space embedding.
Meanwhile, there is some work attempting to use neural structures to encode complex queries.
The transformer structure with specially designed graph hierarchy encoding is proposed by BiQE \cite{kotnis2021answering} for complex queries.
Meanwhile, \citet{DBLP:journals/corr/abs-2302-13114} propose to linearize the computational graph and then use sequence encoders to encode queries.
Newlook \cite{liu2021neural} proposes to use different types of neural networks to iteratively encode complex queries. 
Simultaneously, \cite{amayuelas2022neural} managed to use MLP and MLP-Mixers \cite{tolstikhin2021mlp} to achieve improved performance.  
ENeSy \cite{xu2022neural} uses an entangled method of the neural and symbolic method to conduct query encoding.
QE-GNN \cite{zhu2022neural} uses the message passing over KG to conduct query encoding.
The Mask-and-Reasoning \cite{liu2022mask} proposes to use a knowledge graph transformer, which is a type of graph neural network, to conduct pre-training on the knowledge graph. The pre-trained KG transformer is then fine-tuned to answer complex queries. 

Meanwhile, query decomposition \cite{arakelyan2020complex} is another line of research for answering complex KG queries.
In the query decomposition method, a complex query is first decomposed into atomic queries. Then the probabilities of atomic queries are computed by a neural link predictor.
Recently, \citet{DBLP:journals/corr/abs-2301-08859} propose to use one-hop message passing on query graphs to conduct complex query answering. 
In the inference process, either continuous optimization or beam search is used for finding the answers. 
More recently, SMORE \cite{ren2022smore} is proposed as a framework to train and evaluate large scaled knowledge graph completion and complex query answering models. 
The compositional generalization capability of complex query answering is bench-evaluated by the EFO-1 dataset \cite{wang2021benchmarking}.
On the other hand, ROMA \cite{xi2022reasoning} is proposed to answer complex logic queries on multi-view knowledge graphs. 
Though existing methods can effectively conduct reasoning on entities and their relations on a knowledge graph, they cannot be directly used to reasonably answer numerical complex queries.
Recently, \citet{bai2023complex} propose to use memory-enhanced query encoding for complex query answering on eventuality knowledge graphs. 

Prior research has explored the treatment of numerical attributes in knowledge graphs. 
KBLRN \cite{garcia2017kblrn} proposes to learn the representation of knowledge bases by jointly using latent, relational, and numerical features.
KR-EAR \cite{lin2016knowledge} proposes a knowledge representation model to jointly learn the entity, relations, and attributes in knowledge graphs. 
Numerical value propagation \cite{kotnis2018learning} is proposed to predict the value of entities' numerical attributes. 
Multi-relational attribute propagation \cite{bayram2021node} is then proposed to improve attribute completion by using message passing on a knowledge graph.
However, these methods are not able to answer complex logic queries. 
\citet{DBLP:conf/emnlp/DuanYT21} propose to use the KG embedding methods to train the representations of numerical values over synthetic numerical relations, and this method can is an alternative to the DICE and Sinusoidal for encoding real numbers for NRN. 
Meanwhile, Neural-Num-LP \cite{wang2019differentiable} proposes to learn multi-hop rules involving numerical attributes from a knowledge graph. 
However, it still cannot deal with complex logic queries. 

\vspace{-0.1cm}
\section{Conclusion}

In this work, we proposed the new task of numerical complex attribute query answering (Numerical CQA). 
Meanwhile, we constructed a benchmark dataset based on three public KGs: FB15k, YAGO15k, and DBpedia15k. 
Finally, we proposed a new framework that can conduct numerical attribute value answering. Experiments show that our number encoding network (NRN) with two-phase query encoding can significantly outperform the previous query encoding methods and achieve state-of-the-art Numerical CQA.

\begin{acks}
The authors of this paper are supported by the NSFC Fund (U20B2053) from the NSFC of China, the RIF (R6020-19 and R6021-20), and the GRF (16211520 and 16205322) from RGC of Hong Kong, the MHKJFS (MHP/001/19) from ITC of Hong Kong and the National Key R\&D Program of China (2019YFE0198200) with special thanks to HKMAAC and CUSBLT.
We also thank the UGC Research Matching Grants (RMGS20EG01-D, RMGS20CR11, RMGS20CR12, RMGS20EG19, RMGS20EG21, RMGS23CR05, RMGS23EG08).
\end{acks}

\bibliographystyle{ACM-Reference-Format}
\balance
\bibliography{sample-base}

\newpage
\appendix

\begin{algorithm}
	\caption{Ground General Type} 
	\label{alg:grounding_queries}
	\begin{algorithmic}[h]
	    \Require {$G $} is a knowledge graph.
	    \Function {GroundGeneralType}{$T, v$}
    	    \State {$T$} is an arbitrary node of the computation graph. 
    	    \State {$v$} is an arbitrary knowledge graph vertex
    	    \If {$T.operation = p$}
    	        \State $u \leftarrow \Call{Sample} {\{u| (u, v)\text{is an edge in } G \}} $
    	        \State $RelType \leftarrow \text{type of } (u, v) \text{ in } G$
    	        \If {$v.type = Entity \And u.type = Entity$}
    	            \State $ProjectionType \leftarrow rp$
    	        \ElsIf {$v.type = Entity \And u.type = Values$}
    	            \State $ProjectionType \leftarrow rap$
	            \ElsIf {$v.type = Values \And u.type = Entity$}
	                \State $ProjectionType \leftarrow ap$
    	        \Else
    	            \State $ProjectionType \leftarrow np$
    	        \EndIf
    	        \State $SubQuery \leftarrow \Call{GroundGeneralType}{T.child,u}$  
    	        \State \Return $(ProjectionType, RelType, SubQuery)$
    	    \ElsIf{$T.operation = i$ }
    	        \State {$IntersectionResult = (i) $}
    	        \For {$child  \in T.Children$}
    	            \State $SubQuery \leftarrow\Call{GroundGeneralType}{T.child,v}$
    	            \State {$IntersectionResult.\Call{pushback}{child, v} $}
    	        \EndFor
    	        \State \Return $IntersectionResult$
    	   \ElsIf{$T.operation = u$ }
    	        \State {$UnionResult = (u) $}
    	        \For {$child  \in T.Children$}
        	        \If{$UnionResult.length > 2$}
        	           \State {$v \leftarrow \Call{Sample}{G}$}
    	            \EndIf
    	            \State $SubQuery \leftarrow\Call{GroundGeneralType}{T.child,v}$
    	            
    	            \State {$UnionResult.\Call{pushback}{child, v} $}
    	        \EndFor
    	        \State \Return $UnionResult$
    	    \ElsIf{$T.operation = e$ }
    	        \If{$T.type = Entity$}
    	            \State \Return $(e, T.value)$
    	        \Else
    	            \State \Return $(nv, T.value)$
    	        \EndIf
    	    \EndIf
	    \EndFunction
	\end{algorithmic} 
\end{algorithm}

\section{Sampling Algorithm}
In this section, we introduce the algorithm used for sampling the numerical complex queries from a given knowledge graph. 
The detailed algorithm is described in Algorithm \ref{alg:grounding_queries}.
For a given knowledge $G$ and a general query type $t$, we start with a random node $v$ to reversely find a query that has answer $v$ with the corresponding structure $t$. Basically, this process is conducted in a recursion process. In this recursion, we first look at the last operation in this query. If the operation is projection, we randomly select one of its predecessors $u$ that holds the corresponding relation to $v$ as the answer of its sub-query. Then we call the recursion on node $u$ and the sub-query type of $t$ again. 
Similarly, for intersection and union, we will apply recursion on their sub-queries on the same node $v$. Differently for union, only one of the sub-queries is keeping $u$, while the others will use a randomly selected node, due to the nature of the union operation. 
The recursion will stop when there are no more operations for the current node. 
The specific query types will be determined according to the relation types during the process of sampling projection operation.

\end{document}